\documentclass[conference,10pt]{IEEEtran}
\usepackage{epsfig,rotating,setspace,latexsym,amsmath,epsf,amssymb,amsfonts,bm,theorem,subfigure,epstopdf}
\usepackage{cite,authblk}
\usepackage{bbm}
\usepackage{color}
\usepackage{mathtools}
\usepackage{algorithm}
\usepackage{algpseudocode}
\usepackage{verbatim}
\usepackage{xcolor}

\makeatletter
\newcommand\fs@betterruled{%
  \def\@fs@cfont{\bfseries}\let\@fs@capt\floatc@ruled
  \def\@fs@pre{\vspace*{5pt}\hrule height.8pt depth0pt \kern2pt}%
  \def\@fs@post{\kern2pt\hrule\relax}%
  \def\@fs@mid{\kern2pt\hrule\kern2pt}%
  \let\@fs@iftopcapt\iftrue}
\floatstyle{betterruled}
\restylefloat{algorithm}
\makeatother

\algrenewcommand\algorithmicforall{\textbf{foreach}}
\algrenewcommand\algorithmicindent{0.8em}

\algnewcommand\algorithmicforeach{\textbf{for each}}
\algdef{S}[FOR]{ForEach}[1]{\algorithmicforeach\ #1\ \algorithmicdo}

\IEEEoverridecommandlockouts
\allowdisplaybreaks

\begin{document}

\title{Efficient Semantic Communication Through Transformer-Aided Compression}

\author[1]{Matin Mortaheb}
\author[2]{Mohammad A. (Amir) Khojastepour}
\author[1]{Sennur Ulukus}

\affil[1]{\normalsize University of Maryland, College Park, MD}
\affil[2]{\normalsize NEC Laboratories America, Princeton, NJ}

\maketitle

\begin{abstract}
Transformers, known for their attention mechanisms, have proven highly effective in focusing on critical elements within complex data. This feature can effectively be used to address the time-varying channels in wireless communication systems. In this work, we introduce a channel-aware adaptive framework for semantic communication, where different regions of the image are encoded and compressed based on their semantic content. By employing vision transformers, we interpret the attention mask as a measure of the semantic contents of the patches and dynamically categorize the patches to be compressed at various rates as a function of the instantaneous channel bandwidth. Our method enhances communication efficiency by adapting the encoding resolution to the content's relevance, ensuring that even in highly constrained environments, critical information is preserved. We evaluate the proposed adaptive transmission framework using the TinyImageNet dataset, measuring both reconstruction quality and accuracy. The results demonstrate that our approach maintains high semantic fidelity while optimizing bandwidth, providing an effective solution for transmitting multi-resolution data in limited bandwidth conditions.
\end{abstract}

\section{Introduction}
As we move towards 6G communication systems, semantic communication is emerging as a key enabler of advanced, immersive applications. Unlike traditional communication systems that focus on accurate data transmission, semantic communication \cite{sagduyu2023task, mortaheb2024transformer, zhang2022unified, mortaheb_multi_semantic} prioritizes the meaning and purpose behind the transmitted data. This shift is crucial for supporting next-generation services, such as holographic telepresence \cite{holographicBo} and haptic/tactile communication \cite{qiao2020haptic}, where the goal is not just to reconstruct data but to ensure that the transmitted content fulfills a specific objective in real-time, often with stringent bandwidth and latency constraints. In holographic telepresence \cite{holographicBo}, for example, high-quality 3D holograms must be rendered at the receiver, demanding efficient communication that prioritizes essential information, such as key facial features or body movements. Similarly, in haptic/tactile communication \cite{qiao2020haptic}, the tactile sensations transmitted for remote surgery or virtual reality interactions must be delivered with ultra-low latency and high accuracy to ensure precision.

In this paper, we present an alternative view for semantic communication as described in Section \ref{sec:definition}, where the objective extends beyond simple data reconstruction. In the most common view of semantic communication, the aim is to transmit and reconstruct an image at the receiver, which minimizes the inaccuracies in visual and classification performance. In our view of semantic communication, the transmitted data is encoded such that its reconstruction at the receiver has higher fidelity for the parts that are more relevant to its content, say class label. This means that, parts of the image that are more relevant to the class label, and hence are more important for accurate classification, are encoded at higher rates, while less relevant parts or elements unrelated to the semantic content, such as the background, can be conveyed at a lower resolution and lower rate. Note that, the semantics can also be adapted for other objectives, such as anomaly detection, where the set of image class labels—and consequently the semantics—would change. This approach is sometimes referred to as goal-oriented communication \cite{sagduyu2023task, mortaheb_multi_semantic, yang2023witt, zhang2022unified}.

\begin{figure}[t]
    \centerline{\includegraphics[width=1\linewidth]{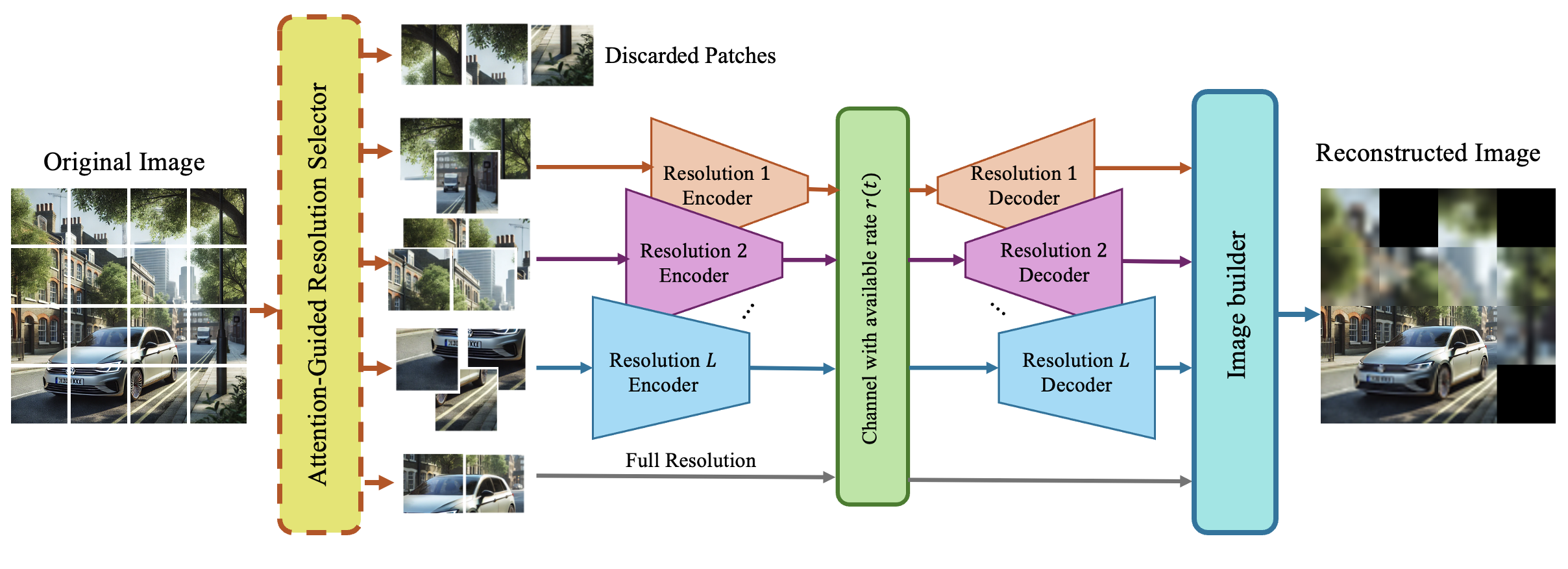}}
    \caption{Channel-aware multi-resolution semantic communication framework.}
    \centering
    \label{fig:framework}
    \vspace*{-0.2cm}
\end{figure}

In our approach, we extend the concept of semantic communication to address the challenges posed by fading channels and varying channel capacities. Specifically, we use separate source and channel coding. Due to the fading, the channel rate will be varying over time. Hence, for each block of the transmission, we will adapt the source encoding rate to the available channel rate. This real-time, adaptive transmission process ensures that the reconstructed image at the receiver maintains high fidelity in its most important regions, while less critical areas can be represented with lower resolution or even left blank. By optimizing which image patches to transmit at varying resolutions, based on their semantic importance, we can adhere to the channel constraints dynamically, achieving better overall performance compared to conventional methods.

In \cite{mortaheb2024transformer}, the concept of using attention scores \cite{vaswani2017attention} in vision transformer (ViT) \cite{dosovitskiy2020image} is introduced for semantic communication. The work \cite{mortaheb2024transformer} generates a binary mask on patches of the image based on the attention scores computed by ViT in order to match the transmission rate to the available channel bitrate and transmit the portion of the image which is more essential. 
Nonetheless, even such a simple binary mask has shown to be very effective in preserving the semantic content of the image. This result reinforces that the attention scores of each patch to the class label are well correlated with the semantic contents of the corresponding patches. In this paper, we explore the use of multi-level quantization of the attention score and introduce multi-resolution encoding of the patches based on their semantic contents as quantified by their respective attention scores to the class label. Our results indicate that this approach boosts the classification accuracy of the reconstructed image at the receiver side. 

Several works \cite{sagduyu2023task, dosovitskiy2020image, mortaheb_multi_semantic} have looked at end-to-end model for semantic communication of images, however, two shortcomings remain to be addressed. First, these works encode the image as a whole without partitioning the image into patches; hence, it is not possible to encode the image patches with different resolutions. Second, the encoding size is static; hence, it is not possible to adapt the encoding rate to the available channel bitrate. As a result, the obtained reconstructed data has lower visual quality related to the semantic content of the image and lower accuracy performance in classification. 

Transformers have also been used in semantic communications \cite{yang2023witt, zhang2022unified, wu2023transformer}. For example, \cite{zhang2022unified} proposes a unified transformer-based model that encodes multimodal data for a variety of tasks, and \cite{wu2023transformer} presents an end-to-end framework for wireless image transmission, leveraging transformers to extract semantics from source images and feedback signals received during each interaction to generate coded symbols effectively. These works use transformers to extract combined semantic information from multimodal data. However, the application of transformers in our work focuses on finding the relevant pieces of image to the semantic content of the image.  

The work \cite{devoto2024adaptive} investigates use of attention scores to pass embeddings from one layer to the next, aiming to reduce computational overhead during transformer model fine-tuning. This approach can be understood as prioritizing the most relevant embeddings for progression through the layers. While this work falls outside the scope of semantic communication, it shares a conceptual similarity with our approach of encoding the most relevant patches for semantic communication. However, unlike our method, no encoding is performed on the embeddings selected for layer-to-layer transmission. In contrast, we employ multi-resolution encoding based on the value of the attention scores. To the best of our knowledge, our framework is the first to employ multi-resolution encoding using ViT attention scores, adapting to varying channel conditions within the context of semantic communications.

Our main contribution is the development of a multi-resolution, end-to-end goal-oriented communication system that leverages a transformer to assess the semantic importance of different image regions, tailored to the specific task at hand; see Fig.~\ref{fig:framework}. We dynamically generate multi-level quantized attention masks based on the real-time available channel rate, and then encode image data at varying resolutions accordingly. This approach ensures that, at the receiver, each part of the image is reconstructed in alignment with its semantic importance. Our experimental results on the TinyImageNet dataset \cite{TinyIV} demonstrate that, under varying channel conditions, our multi-resolution framework significantly improves the performance, yielding a notable increase in task accuracy compared to conventional methods.

\begin{figure*}[!t]
    \centerline{\includegraphics[width=0.75\linewidth]{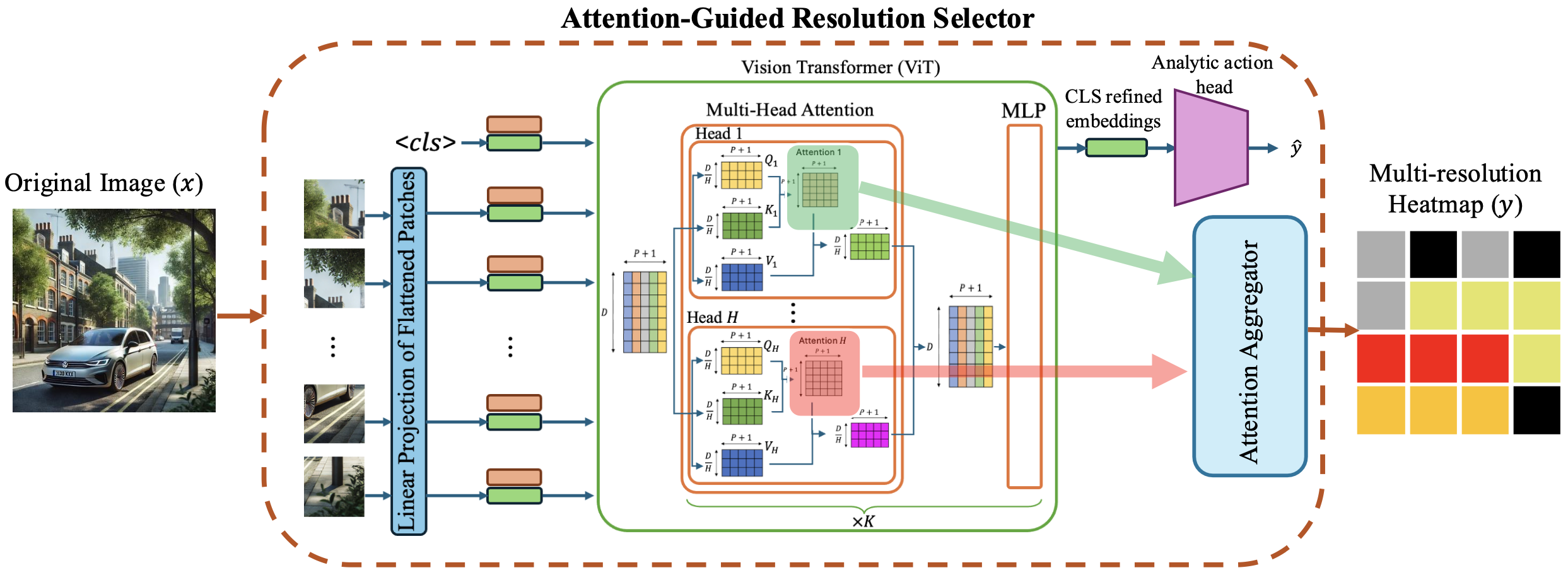}}
    \caption{Attention-guided resolution selector block framework.}
    \centering
    \label{fig:attention_guided_selector}
    \vspace*{-0.2cm}
\end{figure*}

\section{Alternative Views of Semantics and Semantic Communications} \label{sec:definition}
In the following, we formally define semantic meaning in the context of stochastic processes. Let $\mathcal{S}$ denote the context that is defined as the set of all possible semantic classes. For a semantic class $S \in \mathcal{S}$, we observe an instance of a medium (say an image) denoted by $X$ where $X$ is a random variable with distribution $p(X|S)$. The semantic classes follow distribution $p(S)$. The semantic content of $X$ at entropy (or encoding rate) $r$ is defined as $W$ which minimizes $d(X,S; W)$ such that $H(W) \leq r$, where $d(X,S;W)$ defines a distortion measure between $(X,S)$ and $(\hat{X}(W), \hat{S}(W))$, and $\hat{X}(W)$ and $\hat{S}(W)$ are the reconstruction of observation $X$ and its semantic class $S$ based on the semantic content $W$, respectively.

The distortion measure $d(X,S;W)$ may be in the form of linear combination of two disjoint distortion measures, e.g., $d_{linear}(X,S;W) = \alpha_s d_s(S,\hat{S}) + \alpha_x d_x(X,\hat{X})$ for some positive real numbers $\alpha_s$ and $\alpha_x$. Many prior works directly utilize $d_{linear}(X,S;W)$ to extract semantic contents from the data \cite{sagduyu2023task}. For example, $d_{linear}(X,S;W)$ with cross entropy and mean squared error (MSE) as $d_s$ and $d_x$, respectively, is commonly used as a loss function to train, e.g., auto encoders \cite{sagduyu2023task, mortaheb_multi_semantic} for semantic communications. An alternative distortion measure may be defined as $d_{entropy}(X,S;W) = H(S,X|W)$, where minimization of $H(X,S|W)$ corresponds to maximization of $I(X,S;W) = H(X,S) - H(X,S|W)$ as $H(X,S)$ is independent of encoding process and $W$. The maximization of the latter distortion measure may be interpreted as maximizing the information that is in semantic content $W$ about the $(S,X)$ under the constraint of limited entropy (equivalently limited encoding rate), i.e., $H(W) \leq r$.

When the observed medium $X = (X_1, X_2, \ldots, X_P)$ is multidimensional, some dimensions may be more relevant to the the semantic content. For example, consider an image $X$ that is divided into $P$ patches with a class label $S = ``dogs"$. Clearly, the patches which contain the information about the dog are more relevant to the semantic class than the background patches. The same interpretation can be given in the context of random processes, as the distribution for different $X_i$ varies, and the probability distribution $p(X_i|S)$ has different entropy. Consider a case that the decoding of image patches is disjoint, hence, the encoded codeword $W$ is partitioned into $(W_1, W_2, \ldots, W_P)$ where $X_i$ is reconstructed as $\hat{X_i}$ based on $W_i$. Clearly, the sum of the encoding rates for $W_i, 1 \leq i \leq P$ has to be bounded by $H(W) \leq r$. Consider linear distortion measure. The distortion $d_x(X,\hat{X})$ is equal to the sum of the distortion for each patches which depends on the partitioning of the rate $r$ into $\left\{r_i = H(W_i)\right\}_{i=1}^P$. Similarly, the distortion $d_s(S,\hat{S})$ also depends on $W$ which in turn depends on such partitioning of $r$ to $\left\{r_i\right\}_{i=1}^P$. In this paper, we use a suboptimal two step approach, where, first we try to find such partitioning of the rate based on semantic classes using attention scores provided by a vision transformer. Then, we minimize the distortion MSE for each patch $i$ of the image based on the calculated rate budget $r_i$.

\section{System Model and Problem Formulation} \label{sec:prob_formulation}
We address the goal-oriented problem of transmitting an image over a fading channel, where the objective extends beyond just reconstructing the image accurately at the receiver. Our aim is to reconstruct the image in a way that enables the achievement of a specific task, such as image classification, even under varying channel conditions. To accomplish this, we divide the image into patches of a certain size. Each patch is then encoded at the transmitter based on its semantic relevance to the defined goal. At the receiver, the patches are reconstructed according to the resolution received, ensuring that the most important information is preserved while adapting to the channel rate. Our end-to-end system contains three main parts: attention-guided resolution selector, encoder, and decoder.

\subsection{Attention-Guided Resolution Selector}

The attention-guided resolution selector assigns appropriate resolutions for each patch of the image to be encoded with varying rates, depending on their semantic content and available channel rate. Consider the image $X \in \mathcal{X}$ of size $(W, H, C)$, where $W$ and $H$ represent the width and height of the image, respectively, and $C = 3$ is the number of color channels (i.e., RGB). The image $X$ is divided into patches of fixed size $p \times p$ for a total of $P = \frac{WH}{p^2}$ patches. Each patch is mapped by using a linear layer to an embedding of size $d$ that is matched with the ViT embedding size. An additional embedding, called the CLS token, of the same size is appended to the beginning of the patch embeddings for each image. This CLS token is trained and used for classification purposes.

As shown in Fig.~\ref{fig:attention_guided_selector}, the sequence of $P+1$ embeddings, including the CLS and $P$ combined patch embeddings with the corresponding positional embedding is passed through $K$ layers of transformer blocks, each containing a multi-head attention (MHA) unit and a multi-layer perceptron (MLP). The output of the final transformer block produces the output of the vision transformer, consisting of $P+1$ refined embeddings: the first corresponds to the CLS token, and the remaining ones correspond to each of the image patches.

During training, the CLS token's embedding is passed through an analytic action head, which predicts the class $\hat{y}$ based on the task or desired analytic action. For instance, if the task is image classification, the CLS embedding is mapped to an image class while in the case of anomaly detection, the CLS embedding is mapped to either 0 (no anomaly detected) or 1 (anomaly detected). The loss function is then computed using cross-entropy, as defined below, and this loss is used to train the entire ViT, including all MHA units, MLP layers, the analytic action head, and the initial linear layer,
\begin{align}
    \mathcal{L}_{\text{vision transformer}}=\mathcal{L}_{CrossEntropy}(\hat{y},c),
\end{align}
where $c$ is the correct class for the task in hand. The attention matrix of size $(P+1, P+1)$ is computed using the query $Q$ and key $K$ matrices in each attention head, followed by softmax normalization, as
\begin{align}
\text{A}^{(h)}(Q, K) = \text{softmax}\left(\frac{Q^{(h)}K^{{(h)}^T}}{\sqrt{d}}\right).
\end{align} 

Once the vision transformer is trained, the attention matrix of the last transformer block is highly effective in finding the relevancy of each patch to the class label. Specifically, the first row of the attention matrix is a vector of length $1+P$, where the last $P$ values of the vector correspond to the cross attention between CLS and each image patches. The higher the attention value, the higher the importance of the corresponding patch for the desired analytic action. Using the attention matrix for all the heads in the last transformer block, we create a multi-resolution map of size $\frac{W}{p} \times \frac{H}{p}$ that determines the appropriate encoding level for each patch.

\begin{figure}[h]
    \centerline{\includegraphics[width=1.0\linewidth]{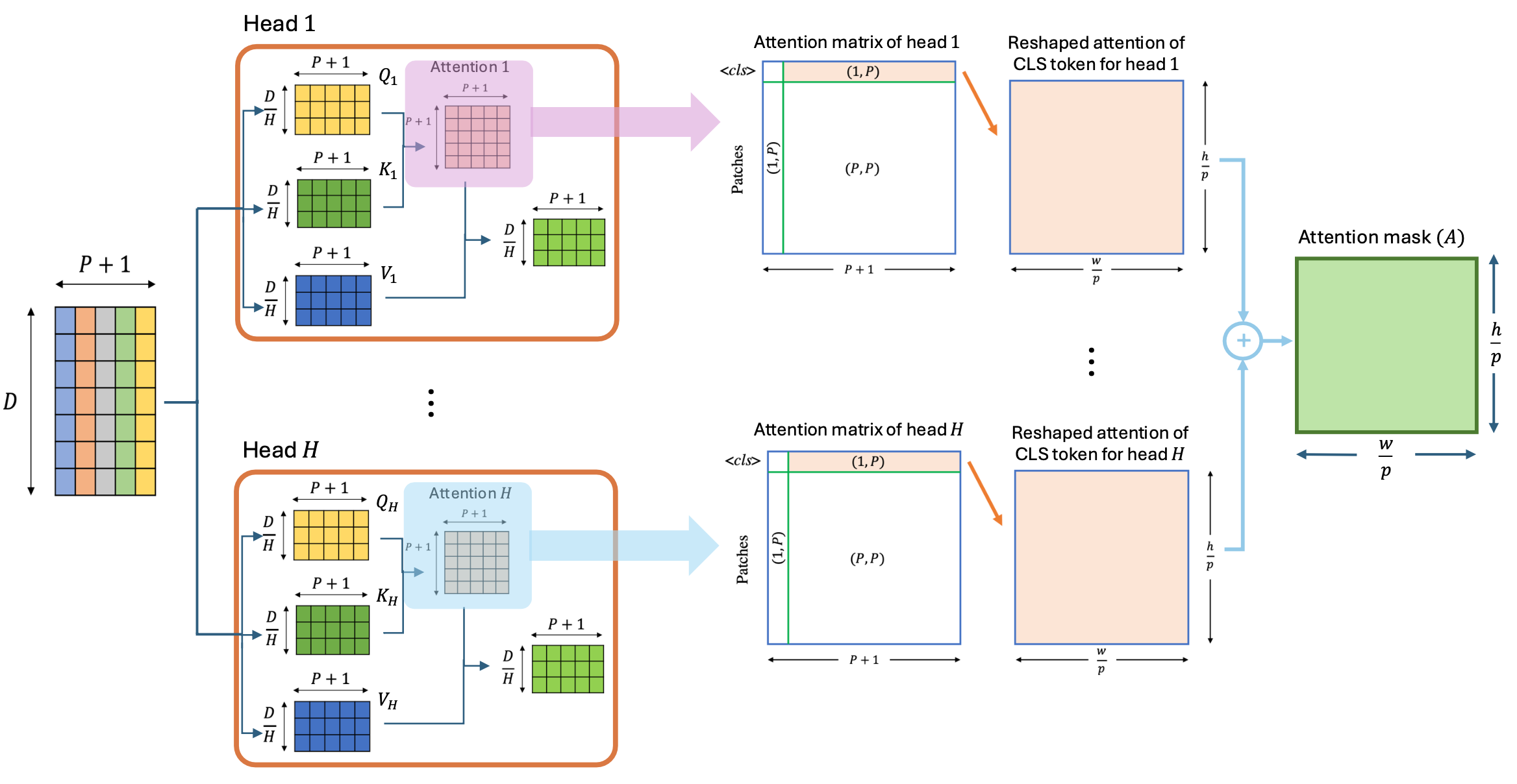}}
    \caption{Attention aggregation block.}
    \centering
    \label{fig:attention_average}
    \vspace*{-0.2cm}
\end{figure}

As shown in Fig.~\ref{fig:attention_average}, we first generate an attention mask of size $\frac{W}{p} \times \frac{H}{p}$ based on the first row of the attention matrix for each head in the last transformer block. Then, we find the average attention mask for all $H$ heads. Finally, we quantize the average attention mask to obtain the multi-resolution map. In particular, we solve an optimization problem to quantize the attention scores into appropriate quantization levels that minimizes the quantization error under two constraints: (i) The encoding rate for the multi-resolution map should not exceed the available channel rate, and (ii) the number of patches that are assigned a nonzero-rate is maximized.

Let the available channel rate be $r$, and assume we have $L$ different resolutions for encoding patches. Algorithm~\ref{alg:resolution_selector_alg} provides a solution to the above optimization problem. Algorithm~\ref{alg:resolution_selector_alg} determines the thresholds for each resolution: First, we check whether the channel rate $r$ is sufficient to encode all patches at the lowest available resolution. If the channel rate is too low to encode all patches at this resolution, we select as many patches as possible until the total bitrate meets the channel rate. If the channel rate exceeds the minimum required to encode all patches at the lowest resolution, we proceed with assigning resolution categories to each patch. We begin by calculating the total sum of the attention scores for all patches. Then, we normalize each attention score by multiplying it by $r/\text{sum(attention scores)}$. After normalization, we apply a function LQ (please see Algorithm~\ref{alg:LQ_UQ}) that maps the attention scores to their nearest lower resolution. At this stage, the total encoding bitrate will be less than the available channel rate, meaning we have not fully utilized the channel. To make use of the remaining bandwidth, we apply a function UQ (please see Algorithm~\ref{alg:LQ_UQ}), which maps the attention scores to their nearest higher resolution. We then upgrade the resolution of patches with attention scores that have the smallest gap to their UQ, as these upgrades require the least additional bandwidth. This process is repeated until the total encoding bitrate matches the channel rate. The overall algorithms for LQ, UQ, and resolution selector block are shown in Algorithm \ref{alg:resolution_selector_alg} and \ref{alg:LQ_UQ}.

\begin{algorithm}[h]
    \caption{Attention-guided resolution selector algorithm}
    \label{alg:resolution_selector_alg}
    \begin{algorithmic}[1]
    \State {\bfseries Input:} Resolution map matrix $A \in \mathbb{R}^{(w/p, h/p)}$, available channel rate $r$
    \State {\bfseries Parameters:} $p$ is a patch size, $P$ total number of patches
    \Statex \hrulefill
    \State Rate mapping: $R = \{0: 0, 1: 12, 2: 24, 3: 48, 4: 196\}$
    \State $A \gets$ Flatten $A$ to a 1D array of size $P$
    \State $A \gets A \times (r / sum(A))$ \Comment{Normalizing the attention}
    \State Obtain $T$ and $Q$ from $R$
    \State $mask \gets$ Zero array of the same shape as $A$
    \State $A\_perm \gets$ Indices of $A$ sorted in descending order
    \If{$r \leq R[1] \times P$}
        \For{$ind = 1$ to $\text{int}(r / R[1])$} $mask[A\_perm[ind]] = 1$ \EndFor
    \Else
        \State $LQ\_A \gets LQ(A,T,Q)$ \Comment{Calculate LQ of $A$}
        \State $UQ\_A \gets UQ(A,T,Q)$ \Comment{Calculate UQ of $A$}
        \State $A[LQ\_A==196] = -Inf$
        \State $LQ\_A[LQ\_A = 0] = R[1]$
        \State $sum\_LQ\_rate = sum(LQ\_A)$
        \While{$sum\_LQ\_rate < r$}
            \State $diff = UQ\_A - A$
            \State $diff\_perm \gets$ $diff$ sorted in ascending order
            \State $LQ\_A[diff\_perm[0]] \gets UQ\_A[diff\_perm[0]]$
            \State $A[diff\_perm[0]] \gets LQ\_A[diff\_perm[0]] + \epsilon$
    
            \If{$UQ\_A[diff\_perm[0]] = 196$}
                \State $A[diff\_perm[0]] = -Inf$
            \EndIf  
            \State $sum\_LQ\_rate = sum(LQ\_A)$
            \State $UQ\_A \gets UQ(A,T,Q)$
        \EndWhile
    \EndIf
    
    \If{$r > R[1] \times P$}
        \State Replace values in $LQ\_attn$: $R[1] \rightarrow 1$, $R[2] \rightarrow 2$, $R[3] \rightarrow 3$, $R[4] \rightarrow 4$
    \Else
        \State $LQ\_attn \gets mask$
    \EndIf
    
    \State $mask\_attn \gets$ Reshape $LQ\_attn$ to a $w/p \times h/p$ array
    \State \Return $mask\_attn$
    \end{algorithmic}
\end{algorithm}

\begin{algorithm}[h]
\caption{Lower and Upper Quantization Functions}
\label{alg:LQ_UQ}
\begin{algorithmic}[1]
\Require Array $y$, Threshold levels $T = [t_1, t_2, \dots, t_n]$, Quantized values $Q = [q_0, q_1, \dots, q_{n+1}]$
\Statex \hrulefill
\Function{LQ}{$y, T, Q$} \Comment{Lower Quantization}
    \State $x \gets y.copy()$
    \For{$i = 0$ to $\text{len}(x) - 1$}
        \State Find the highest index $j$ such that $x[i] < T[j]$
        \State $x[i] \gets Q[j]$
    \EndFor
    \State \Return $x$
\EndFunction

\Function{UQ}{$y, T, Q$} \Comment{Upper Quantization}
    \State $x \gets y.copy()$
    \For{$i = 0$ to $\text{len}(x) - 1$}
        \State Find the lowest index $j$ such that $x[i] > T[j]$
        \State $x[i] \gets Q[j + 1]$
    \EndFor
    \State \Return $x$
\EndFunction

\end{algorithmic}
\end{algorithm}

\subsection{Encoder and Decoder}
The entries of the resolution map indicate the resolutions in which patches are encoded. We train encoder-decoder pairs for each of the resolutions. Each resolution is assigned an encoding size out of a set of possible quantized encoding sizes $\{b_i, 1 \leq i \leq L\}$. The higher the resolution, the higher the assigned rate. For an encoding size $b_i$, the encoder  takes an image patch $X_j, j \in \{1, 2, \ldots, P\}$ as an input and produces an embedding of size $b_i$ and the decoder takes this embedding and produces $\hat{X}_j$. The resolution map is designed to satisfy the available channel bitrate constraint, and relying on error-free communication not exceeding the available bitrate, we train the encoder-decoder for all resolutions as follows. Let $\hat{X}$ denote the reconstructed image for all the patches $\hat{X}_j$ at the encoding sizes given by resolution map for the image $X$, the following loss function can be used for the training,
\begin{align}
    \mathcal{L}_{Encoder-Decoder} = \mathcal{L}_{MSE}(X,\hat{X}).
\end{align}

However, to design a robust framework that is independent of the analytic action function, we train the encoder-decoder pairs independent of the analytic action function. In this case, we train the encoder-decoder pair for each rate on all image patches available for training using the MSE loss function on individual patch $X_j, j \in \{1, 2, \ldots, P\}$ given by,
\begin{align}
    \mathcal{L}_{Encoder-Decoder} = \mathcal{L}_{MSE}(X_j,\hat{X_j}).
\end{align}
In this case, the encoder-decoder pairs for different resolutions are trained independently. Hence, a key advantage of this approach is that the encoder-decoder pairs do not need to be retrained when the semantic meaning or communication goal changes as imposed by different analytic action functions, i.e., the resolution selector retains the semantic meaning by properly incorporating the analytic action function in generation of the resolution map as the communication goal or task changes.

\section{Experimental Results} \label{sec:Exp_section}
We evaluate the performance of our proposed method in terms of reconstruction and accuracy performances.

\subsection{Dataset Specifications}
We use the TinyImageNet dataset \cite{TinyIV}, a simplified version of the larger ImageNet dataset \cite{imagenet}, to train the encoder-decoder pairs. TinyImageNet consists of 200 classes, with each class containing 500 training images, 50 validation images, and 50 test images. In total, the dataset includes 100,000 training images, 10,000 validation images, and 10,000 test images. All images are originally in RGB format with a resolution of $64\times64$ pixels. For our application, we scale all these images up to a resolution of $320\times480$ pixels. The transformer encoder we use has been pre-trained on the full ImageNet dataset \cite{imagenet}, providing a strong initialization for fine-tuning on any goal defined later such as image classification or anomaly detection.

\subsection{Model and Hyperparameters}
\begin{figure}[t]
    \centerline{\includegraphics[width=1\linewidth]{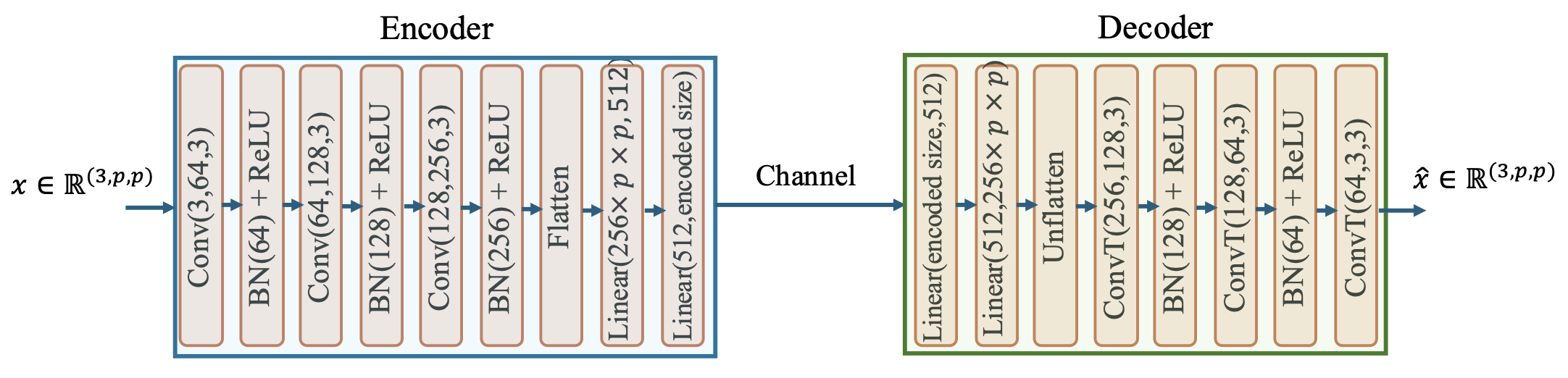}}
    \caption{Encoder and Decoder structure for different resolutions.}
    \centering
    \label{fig:encoderdecoder_model}
    \vspace*{-0.2cm}
\end{figure}

Our proposed framework consists of two primary components: the encoder-decoder pairs for each resolution, and the ViT encoder for deriving the attention scores. In the ViT part, we utilize the DINO \cite{DINO} model as a transformer encoder, which is pre-trained through a self-supervised approach on the ImageNet dataset. DINO contains three variants: \emph{Tiny-ViT}, \emph{Small-ViT}, and \emph{ViT}. For our framework, we select \emph{Small-ViT} with a patch size of $p = 8$. This model has an embedding dimension of $D = 384$, and includes $K = 12$ transformer blocks, each with $H = 6$ attention heads. For the encoder-decoder pair, we have designed a deep neural network (DNN) model that features a sequence of convolutional layers. The detailed structure of this encoder-decoder model is illustrated in Fig.~\ref{fig:encoderdecoder_model}. We use Adam optimizer with learning rate $\eta = 2\times10^{-5}$ to individually train the parameters of both encoder and decoder. The batch size is set to 32. To calculate the accuracy, we use a pre-trained classifier trained on TinyImageNet dataset. 

\subsection{Results and Analysis}
Our framework consists of five resolutions with encoding sizes of $L = 0, 12, 24, 48, 198$ bytes. The first resolution discards the patch encoding prior to transmission to the receiver, while the last resolution transmits the entire image $(3 \times 8 \times 8)$ bytes without further encoding. The three intermediate resolutions involve an encoding and decoding process. Each patch is passed through an encoder to map it to the predefined size of the selected resolution, and the decoder then reconstructs the patch back to its original image size. 

\begin{figure}[t]
    \centerline{\includegraphics[width=0.85\linewidth]{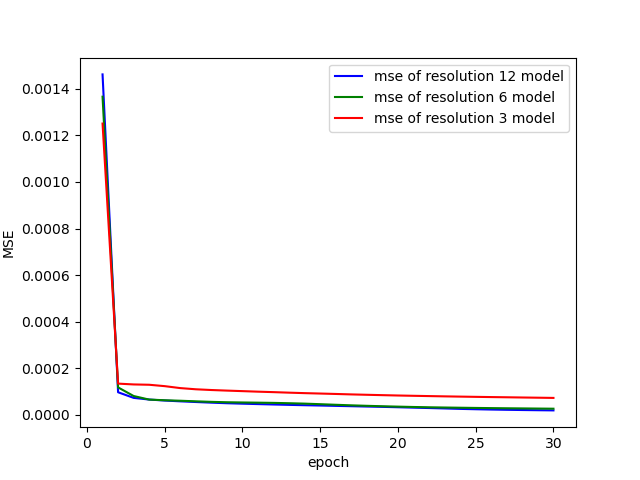}}
    \caption{Reconstruction result for three medium resolutions.}
    \centering
    \label{fig:reconstruction_performance}
    \vspace*{-0.3cm}
\end{figure}

\begin{figure}[t]
    \centerline{\includegraphics[width=0.85\linewidth]{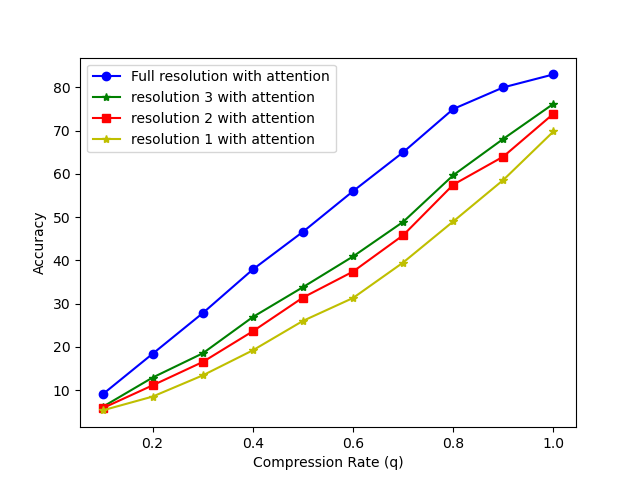}}
    \caption{Accuracy result for three medium resolutions.}
    \centering
    \label{fig:acc_performance}
    \vspace*{-0.2cm}
\end{figure}

The encoder-decoder pairs for each of the three intermediate resolutions are trained on the entire TinyImageNet dataset. Fig.~\ref{fig:reconstruction_performance} and Fig.~\ref{fig:acc_performance} display the reconstruction quality on the training data and the accuracy on the test dataset, respectively. The reconstruction results indicate that the encoder-decoder pairs are well-trained, and the accuracy results illustrate the performance of the encoder-decoder system across different compression rates. For instance, with a compression rate $q = 0.5$, half of the patches are selected based on the attention scores, then transmitted to the receiver for decoding and image reconstruction. All selected patches are encoded with a single resolution. As expected, the accuracy results demonstrate that higher encoded sizes (better resolution) significantly improve accuracy across varying compression levels.

\begin{figure}[t]
    \centerline{\includegraphics[width=0.85\linewidth]{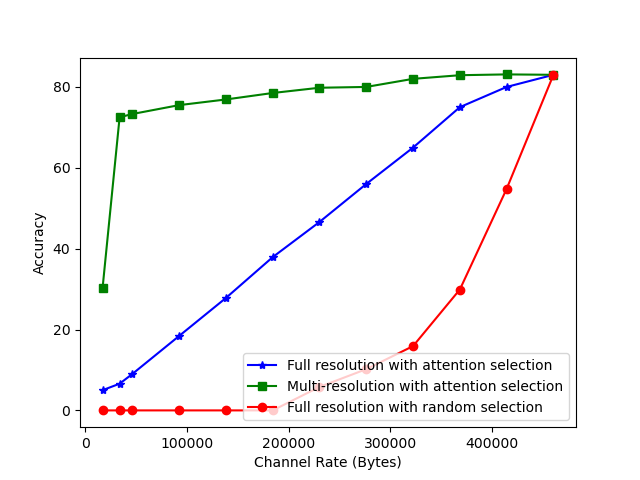}}
    \caption{Accuracy result for adaptive multi-resolution semantic communication framework in various channel rates.}
    \centering
    \label{fig:acc_performance_dynamic}
    \vspace*{-0.3cm}
\end{figure}

\begin{figure}[t]
    \centerline{\includegraphics[width=0.85\linewidth]{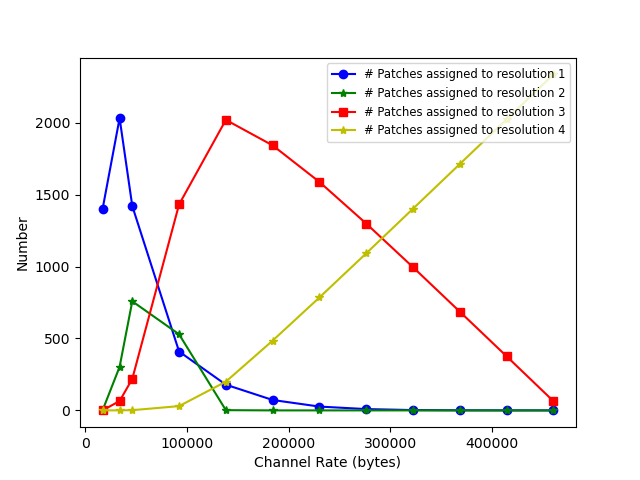}}
    \caption{Resolution assignment to patches in different channel rate.}
    \centering
    \label{fig:res_assignment_performance}
    \vspace*{-0.2cm}
\end{figure}

Next, we evaluate the accuracy performance when using multiple resolutions (using Algorithm~\ref{alg:resolution_selector_alg}) based on the available channel rate. We compare this multi-resolution approach to the case where a single fixed resolution is used for transmission. The results show that employing a multi-resolution framework achieves higher accuracy at higher compression rates (i.e., lower available channel rates). As expected, transmitting the most valuable patches at higher resolutions, while sending lower-importance patches at lower resolutions, helps the pretrained classifier achieve higher accuracy even at lower rates. At higher rates, most patches are transmitted at higher resolutions, ultimately matching the performance of full-resolution transmission. The accuracy performance for the multi-resolution framework is shown in Fig.~\ref{fig:acc_performance_dynamic}. Fig.~\ref{fig:res_assignment_performance} also illustrates the distribution of image patches transmitted at different resolutions across various channel rates. At lower rates, the resolution selector primarily assigns patches to lower resolution embedding sizes to meet the channel constraints while preserving the semantic content. In contrast, at higher rates, most patches are assigned to higher resolutions to fully utilize the channel capacity.

\bibliographystyle{unsrt}
\bibliography{reference}
\end{document}